\documentclass[11pt]{article}
\usepackage{naacl2021} 
\usepackage{naacl2021}  
\usepackage{times}
\usepackage{graphicx}
\usepackage{wrapfig}
\usepackage{dsfont}
\usepackage{latexsym}
\usepackage[linesnumbered,algoruled,boxed,noend]{algorithm2e}
\usepackage{amssymb}
\usepackage{amsmath}
\usepackage{url} 
\SetKwInOut{Parameter}{parameter}
\usepackage{enumitem}
\usepackage{mathtools}
\usepackage{cleveref}
\usepackage{multirow}
\usepackage{hhline}
\usepackage[export]{adjustbox}
\usepackage{booktabs,array}
\usepackage{amsmath, bm}
\usepackage{color, colortbl}
\usepackage{xcolor}
\usepackage{resizegather}
\SetKwInput{KwInput}{Input}
\SetKwInput{KwOutput}{Output}
\newcolumntype{P}[1]{>{\centering\arraybackslash}p{#1}}
\usepackage{booktabs,makecell,tabularx} 
\setitemize{noitemsep,topsep=0pt,parsep=0pt,partopsep=0pt}
\let\oldnl\nl
\definecolor{Gray}{gray}{0.9}
\definecolor{LightCyan}{rgb}{0.88,1,1}
\newcommand{\nonl}{\renewcommand{\nl}{\let\nl\oldnl}}

\usepackage{pifont}
\newcommand{\cmark}{\ding{51}}%
\usepackage{epigraph}

\usepackage{etoolbox}
\patchcmd{\epigraph}{\@epitext{#1}}{\itshape\@epitext{#1}}{}{}

\newlength{\Width}%
\newlength{\DepthReference}
\newlength{\HeightReference}
\newcommand{\MyColorBox}[2][red]%
{%
    \settowidth{\Width}{#2}%
    \colorbox{#1}%
    {%
        \raisebox{-\DepthReference}%
        {%
                \parbox[b][\HeightReference+\DepthReference][c]{\Width}{\centering#2}%
        }%
    }%
}
\begin{document}

\title{\vspace*{-0.5in}
{{\small \hfill NAACL 2021}\\
\vspace*{.25in}}
Differentiable Open-Ended Commonsense Reasoning}

\author{Bill Yuchen Lin$^1$\thanks{~~The work was mainly done during Bill Yuchen Lin's internship at Google Research.},~
Haitian Sun$^2$, Bhuwan Dhingra$^2$, \\ 
\textbf{Manzil Zaheer$^2$, Xiang Ren$^1$, William W. Cohen$^2$}\\
$^1$ University of Southern California\\
$^2$ Google Research\\
\texttt{\{yuchen.lin,xiangren\}@usc.edu}\\
\texttt{\{haitiansun,bdhingra,manzilzaheer,wcohen\}@google.com}
}


\maketitle
\begin{abstract}
Current commonsense reasoning research  focuses on developing models that use commonsense knowledge to answer \textit{multiple-choice} questions.
However, systems designed to answer multiple-choice questions may not be useful in applications that do not provide a small list of candidate answers to choose from.
As a step towards making commonsense reasoning research more realistic and useful, 
we propose to study \textit{open-ended commonsense reasoning} ({OpenCSR}) --- the task of answering a commonsense question \textit{without} any pre-defined choices --- using as a resource only a knowledge corpus of commonsense facts written in natural language.
OpenCSR is challenging due to a  large decision space, and because many
questions require \textit{implicit} multi-hop reasoning.
As an approach to OpenCSR, 
we propose \textsc{DrFact}, an efficient \underline{D}ifferentiable model for multi-hop \underline{R}easoning over knowledge \underline{Fact}s.
To evaluate OpenCSR methods, 
we adapt three popular multiple-choice datasets, and collect \textit{multiple} new answers to each test question via crowd-sourcing.
Experiments show that \textsc{DrFact} outperforms strong baseline methods by a large margin.%
\footnote{Our code and data are available at the project website --- \url{https://open-csr.github.io/}. The human annotations were collected by the USC-INK group.}



\end{abstract}

\section{Introduction}\label{sec:intro}
The conventional task setting for most current commonsense reasoning research is \textit{multiple-choice} question answering (QA) --- i.e., given a question and a small set of pre-defined answer choices, models are required to determine which of the candidate choices best answers the question.
Existing commonsense reasoning models usually work by scoring a question-candidate pair~\cite{kagnet-emnlp19, lv2020graph, feng2020scalable}. 
Hence, even an accurate multiple-choice QA model cannot be directly applied in practical applications where answer candidates are not provided
(e.g., answering a question asked on a search engine, or during conversation with a chat-bot).

\begin{figure}
	\centering
	\includegraphics[width=1\linewidth]{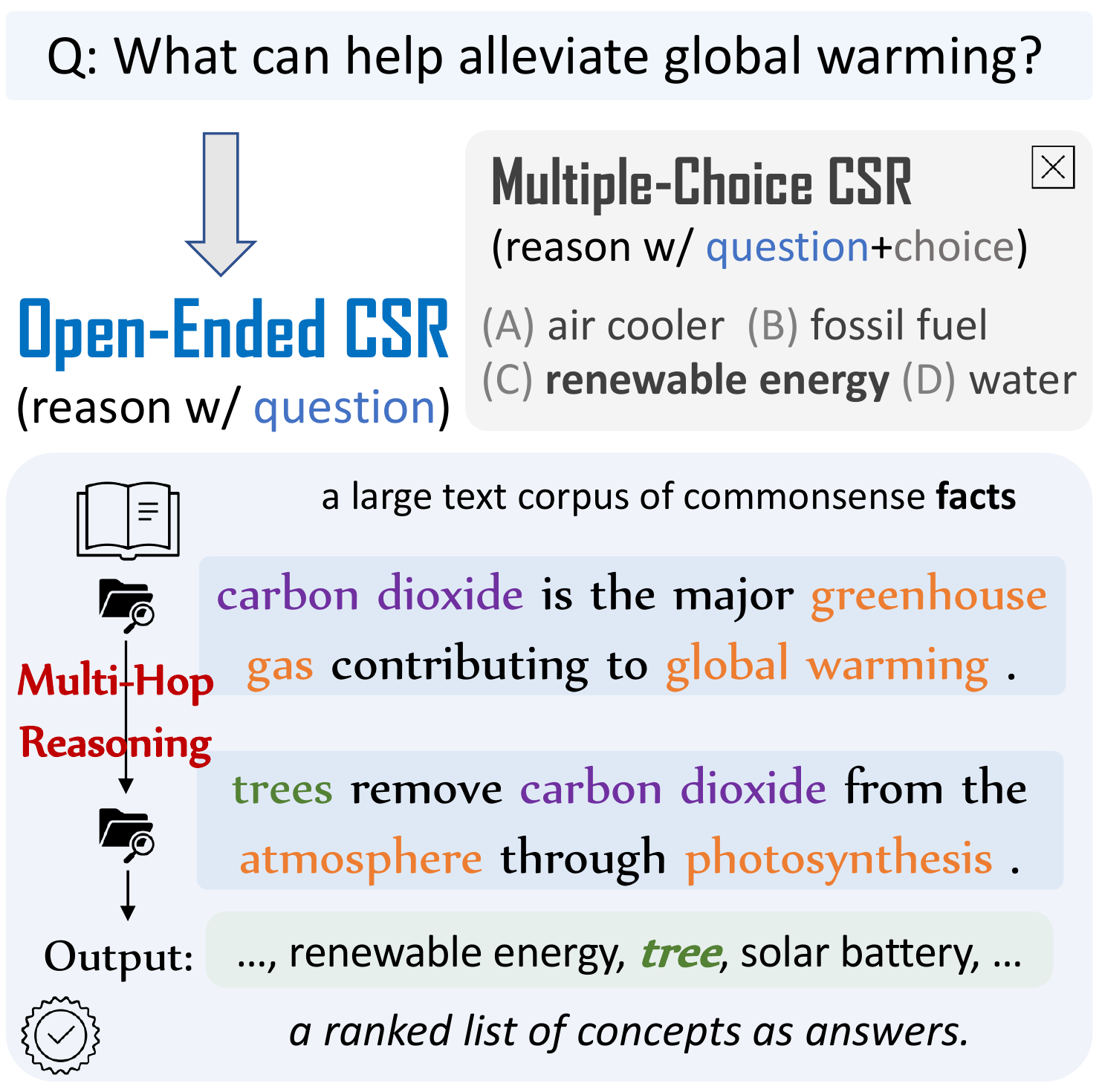}
	\caption{{We study the task of open-ended commonsense reasoning (OpenCSR), where answer candidates are not provided (as in a multiple-choice setting). Given a question, a reasoner uses \textit{multi-hop reasoning} over a knowledge corpus of facts, and outputs a ranked list of concepts from the corpus.}}
	\label{fig:opencsr}
	\vspace{-1em}
\end{figure}

Because we seek to advance commonsense reasoning towards practical applications, 
we propose to study 
{\textbf{{open-ended commonsense reasoning}}} ({OpenCSR}), where answers
are 
generated efficiently, 
rather than selected from a small list of candidates (see Figure~\ref{fig:opencsr}).
As a step toward this, here we explore a setting where the model produces a ranked list of answers from a large question-independent set of candidate concepts that are extracted offline from a corpus of common-sense facts written in natural language.

The OpenCSR task is inherently challenging. 
One problem is that for many questions, finding an answer requires reasoning over two or more natural-language facts from a corpus.  
In the multiple-choice QA setting, as the set of candidates is small,
we can pair a question with an answer, and use the combination to retrieve relevant facts and then reason with them.
In the open-ended setting, this is impractical: instead one needs to retrieve facts
from the corpus using the question alone. 
In this respect, OpenCSR is similar to multi-hop factoid QA about named entities, e.g.~as done for HotpotQA~\cite{yang2018hotpotqa}.

However, the underlying reasoning chains of most multi-hop factoid QA datasets are relatively \textit{clear} and \textit{context-independent}, and are thus easier to infer.
Commonsense questions, in contrast, exhibit more variable types of reasoning, and the relationship between a question and the reasoning to answer the question is often unclear. 
(For example, a factoid question like ``\textit{who starred in a movie directed by Bradley Cooper?}'' {clearly} suggests following a \textit{directed-by} relationship and then a \textit{starred-in} relationship, while the underlying reasoning chains of a question like ``what can help alleviate global warming?'' is relatively implicit from the question.)
Furthermore, 
annotations are not available to identify which facts are needed in the latent reasoning chains that lead to an answer --- the only supervision is a set of questions and their answers.
We discuss the formulation of OpenCSR and its  challenges further in Section~\ref{sec:problem}.

As shown in Fig.~\ref{fig:opencsr}, another challenge is that
many commonsense questions require reasoning about facts that link several concepts together.  
E.g., the fact ``{trees remove carbon dioxide from the atmosphere through photosynthesis}'' cannot be easily decomposed into pairwise relationships between ``trees'', ``carbon dioxide'', ``the atmosphere'', and ``photosynthesis'', which makes it more difficult to store in a knowledge graph (KG).
However, such facts have been collected as sentences in common-sense corpora, e.g., Generics\-KB~\cite{bhakthavatsalam2020genericskb}.
This motivates the question: {how can we conduct \textit{multi-hop} reasoning over such a knowledge corpus, similar to the way multi-hop reasoning methods traverse a KG?} 
Moreover, can we achieve this in a \textit{differentiable} way, to support end-to-end learning?

To address this question, we extend work by~\citeauthor{seo2019real} (2019) and~\citeauthor{drkit} (2020),
and propose an efficient, differentiable multi-hop reasoning method for OpenCSR,
named \textsc{DrFact} (for \underline{D}ifferentiable \underline{R}easoning over \underline{Fact}s).
Specifically, we formulate multi-hop reasoning over a corpus as an iterative process of differentiable \textit{fact-following} operations over a hypergraph. 
We first encode all fact sentences within the corpus as \textit{dense} vectors to form a \textit{neural} fact index, such that a fast retrieval can be done via maximum inner product search (MIPS).
This dense representation is supplemented by  
a \textit{sparse} fact-to-fact matrix to store  \textit{symbolic} links between facts (i.e., a pair of facts are linked if they share common concepts).
\textsc{DrFact} thus merges both neural and symbolic aspects of the relationships between facts to model reasoning
in an end-to-end differentiable framework (Section~\ref{sec:method}).

To evaluate OpenCSR methods, we construct new OpenCSR datasets by adapting three existing multiple-choice QA datasets: 
QASC~\cite{khot2020qasc}, OBQA~\cite{mihaylov2018can}, 
and ARC~\cite{clark2018think}.
Note that unlike \textit{factoid} questions that usually have a \textit{single} correct answer, 
open-ended commonsense questions can have \textit{multiple} correct answers.
Thus, we collect a collection of new answers for each test question by crowd-sourcing human annotations.
We compare with several strong baseline methods and show that our proposed \textsc{DrFact} outperforms them by a large margin. Overall \textsc{DrFact}  gives an 4.6\%  
absolute improvement in Hit@100 accuracy over DPR~\cite{dpr}, a state-of-the-art text retriever for QA, and 3.2\% over DrKIT~\cite{drkit}, a strong baseline for entity-centric multi-hop reasoning.
With a relatively more expensive re-ranking module, the gap between \textsc{DrFact} and others is even larger.
(Sec.~\ref{sec:exp})





\section{Related Work}\label{sec:rel_work}

\textbf{Commonsense Reasoning.}
Many recent commonsense-reasoning (CSR) methods focus on multiple-choice QA.
For example, {KagNet}~\cite{kagnet-emnlp19} and {MHGRN}~\cite{feng2020scalable} use an external commonsense knowledge graph 
as structural priors to individually score each choice.
These methods, though powerful in determining the best choice for a multi-choice question, are less realistic for practical applications where answer candidates are typically not available. 
UnifiedQA~\cite{khashabi2020unifiedqa} and other closed-book QA models~\cite{roberts2020much}  generate answers to questions by fine-tuning a text-to-text transformer such as BART~\cite{lewis2019bart} or T5~\cite{t5}, but a disadvantage of closed-book QA models is that they do not provide intermediate explanations for their answers, i.e., the supporting facts,
which makes them less trustworthy in downstream applications.
Although closed-book models exist that are augmented with an additional retrieval module~\cite{lewis2020retrieval}, these models mainly work for single-hop reasoning.





\smallskip
\noindent
\textbf{QA over KGs or Text.}
A conventional source of commonsense knowledge is triple-based symbolic commonsense knowledge graphs (CSKGs) such as ConceptNet~\cite{Speer2017ConceptNet5A}. 
However, the binary relations in CSKGs greatly limit the types of the knowledge that can be encoded.  
Here, instead of a KB, we use a corpus of generic sentences about commonsense facts, in particular  GenericsKB~\cite{bhakthavatsalam2020genericskb}.  The advantage of this approach is that text can represent more complex commonsense knowledge, including facts that relate three or more concepts.
Formalized in this way,
OpenCSR is a question answering task requiring (possibly) iterative retrieval, similar to other open-domain QA tasks~\cite{chen2017reading} such as HotpotQA~\cite{yang2018hotpotqa} and Natural Questions~\cite{kwiatkowski2019natural}.
As noted above, however, the surface of commonsense questions in OpenCSR have fewer hints about kinds of multi-hop reasoning required to answer them than the factoid questions in open-domain QA, resulting in a particularly challenging reasoning problem (see Sec.~\ref{sec:problem}).

\smallskip
\noindent
\textbf{Multi-Hop Reasoning.}
Many recent models for open-domain QA tackle multi-hop reasoning through iterative retrieval, e.g., GRAFT-Net~\cite{sun2018open}, MUPPET~\cite{feldman-el-yaniv-2019-multi},  PullNet~\cite{sun2019pullnet}, and GoldEn~\cite{qi2019answering}.
These models, however, are \textit{not} end-to-end differentiable and thus tend to have slower inference speed,
which is a limitation shared by many other works using reading comprehension for multi-step QA~\cite{das2019multi, lee2019latent}.  As another approach,
Neural Query Language~\cite{Cohen2020Scalable} designs 
 differentiable multi-hop entity-following templates for reasoning over a compactly stored symbolic KG, but this KG
 is limited to {binary} relations between entities from an {explicitly} enumerated set.

\smallskip
\noindent
\textbf{DrKIT}~\cite{drkit} is the most similar work to our \textsc{DrFact}, as it also supports multi-hop reasoning over a corpus.  Unlike \textsc{DrFact}, DrKIT is designed for entity-centric reasoning.  DrKIT begins with an entity-linked corpus, and  computes both sparse and dense indices of \textit{entity mentions} (i.e., linked named-entity spans).  
DrKIT's fundamental reasoning operation is to ``hop'' from one weighted set of $X$ entities to another, by 1) finding mentions of new entities $x'$ that are related to some entity in $X$, guided by the indices, and then 2) aggregating these mentions to produce a new weighted set of entities.  
DrKIT's operations are differentiable, and by learning to construct appropriate queries to the indices, it can be trained to answer multi-hop entity-related questions. 

Prior to our work DrKIT been applied only on \textit{factoid} questions about named entities.  
In CSR, the concepts that drive reasoning are generally less precise than entities,  harder to disambiguate in context, and are also much more densely connected, so it is unclear to what extent DrKIT would be effective.  We present here novel results using DrKIT on OpenCSR tasks, and show experimentally that our new approach, \textsc{DrFact}, improves over DrKIT.  \textsc{DrFact} mainly differs from DrKIT in that its reasoning process learns to ``hop'' from one fact to another, rather than from one entity to another, thus effectively using the full information from a fact for multi-hop reasoning.



\section{Open-Ended Commonsense Reasoning}
\label{sec:problem}



\begin{figure*}[th]
	\centering
	\includegraphics[width=1\linewidth]{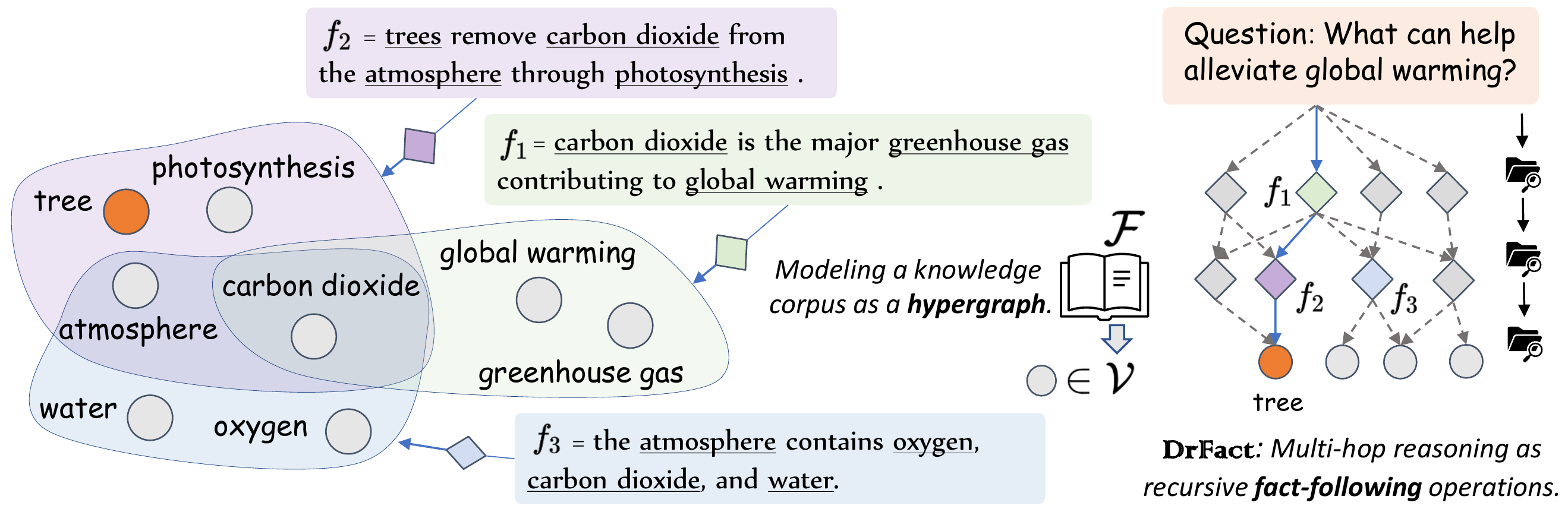}
	\caption{\textbf{A motivating example of how  DrFact works for OpenCSR.} We model the knowledge corpus as a hypergraph consisting of \textit{concepts} in $\mathcal{V}$ as \textit{nodes} and \textit{facts} in  $\mathcal{F}$ as \textit{hyperedges}. Then, we develop a differentiable reasoning method, DrFact, to perform \textit{multi-hop reasoning} via ~\textit{fact-following} operations (e.g., $f_1 \rightarrow f_2$). }
	\label{fig:hypercskg}
\end{figure*}

\smallskip
\noindent
\textbf{Task Formulation.}
We denote a \textbf{corpus} of knowledge facts as $\mathcal{F}$, and use $\mathcal{V}$ to denote a {vocabulary} of \textbf{concepts}; both are sets consisting of unique elements.
A \textbf{fact} $f_i\in \mathcal{F}$ is a sentence that describes generic commonsense knowledge, such as ``{\textit{trees} remove \textit{carbon dioxide} from the \textit{atmosphere} through \textit{photosynthesis}}.''
A \textbf{concept} $c_j\in \mathcal{V}$ is a noun or base noun phrase mentioned frequently in these facts (e.g.,  `tree' and `carbon dioxide').  Concepts are considered identical if their surface forms are the same (after lemmatization).
Given only a \textbf{question} $q$ (e.g., ``\textit{what can help alleviate global warming?}''),
 an open-ended commonsense reasoner is supposed to \textbf{answer} it by returning {a weighted set of concepts}, such as \{($a_1$=\textit{`renewable energy'}, $w_1$), ($a_2$=\textit{`tree'}, $w_2$),
 \dots \}, where $w_i\in \mathbb{R}$ is the weight of the predicted concept $a_i\in \mathcal{V}$. 

To learn interpretable, trustworthy reasoning models, 
it is expected that models can output intermediate results that justify the reasoning process 
--- i.e., the supporting facts from $\mathcal{F}$. 
E.g., an \textbf{explanation} for `tree' to be an answer to the question above can be the combination of two facts: $f_1$ = ``{carbon dioxide} is the major ...'' and $f_2$ = ``{trees} remove ...'', as shown in Figure~\ref{fig:opencsr}.

\smallskip
\noindent
\textbf{Implicit Multi-Hop Structures.}
Commonsense questions 
(i.e., questions that need commonsense knowledge to reason) 
contrast with better-studied multi-hop factoid QA datasets, e.g., HotpotQA~\cite{yang2018hotpotqa}, 
which primarily focus on querying about \textit{evident relations between named entities}.
For example, an example multi-hop factoid question can be ``which team does the player named 2015 Diamond Head Classic's MVP play for?''
Its query structure is relatively clear and \textit{self-evident} from the question itself: in this case the reasoning process can be decomposed into $q_1$ = ``the player named 2015 DHC's MVP'' and $q_2$ = ``which team does $q_1.\operatorname{answer}$ play for''.
%

The reasoning required to answer commonsense questions is usually more \textit{implicit} and relatively unclear.
Consider the previous example in Fig.~\ref{fig:opencsr},
$q$ = `what can help alleviate global warming?' can be decomposed by $q_1$ = ``what contributes to global warming'' and $q_2$ = ``what removes $q_1.\operatorname{answer}$ from the atmosphere'' --- but many other decompositions are also plausible.
In addition, unlike HotpotQA, we assume that we have \textit{no ground-truth justifications} for training, which makes OpenCSR even more challenging.

\section{{DrFact}: An Efficient Approach for Differentiable Reasoning over Facts} 
\label{sec:method}

In this section we present \textsc{DrFact}, a  model for multi-hop reasoning over facts.
More implementation details are in Appendix~\ref{sec:impl}.





\subsection{Overview}
In \textsc{DrFact}, we propose to model reasoning as traversing a \textit{hypergraph}, where each \textit{hyperedge} corresponds to a fact in $\mathcal{F}$, and connects the concepts in $\mathcal{V}$ that are mentioned in that fact.  This is  shown in Figure~\ref{fig:hypercskg}.
Notice that a fact, as a hyperedge, connects multiple concepts that are mentioned, while the textual form of the fact maintains the contextual information of the original natural language statement, and hence we do not assume a \textit{fixed} set of relations.

Given such a hypergraph,
our open-ended reasoning model will traverse the hypergraph starting from the question (concepts) and finally arrive at a set of concept nodes by following multiple hyperedges (facts).
A probabilistic view of this process over $T$ hops is:
{
\begin{equation*}
\resizebox{0.97\hsize}{!}{$P(c \mid q) = 
P (c \mid  q, F_T )  \prod^T_{t=1}  P (F_t \mid q, F_{t-1})  P \left(F_0 \mid q\right)$}
\end{equation*} 
}%


Intuitively, we want to model the distribution of a concept $c\in \mathcal{V}$ being an answer to a question $q$ as $P(c\mid q)$.
This answering process can be seen as a process of multiple iterations of ``fact-following,'' or moving from one fact to another based on shared concepts, and finally moving from facts to concepts.
We use $F_t$ to represent a weighted set of retrieved facts at the hop $t$, and $F_0$ for the initial facts below.
Then, given the question and the current retrieved facts, we iteratively retrieve the facts for the next hop. Finally, we score a concept using  retrieved facts.


\begin{figure*}[th]
	\centering
	\includegraphics[width=1\linewidth]{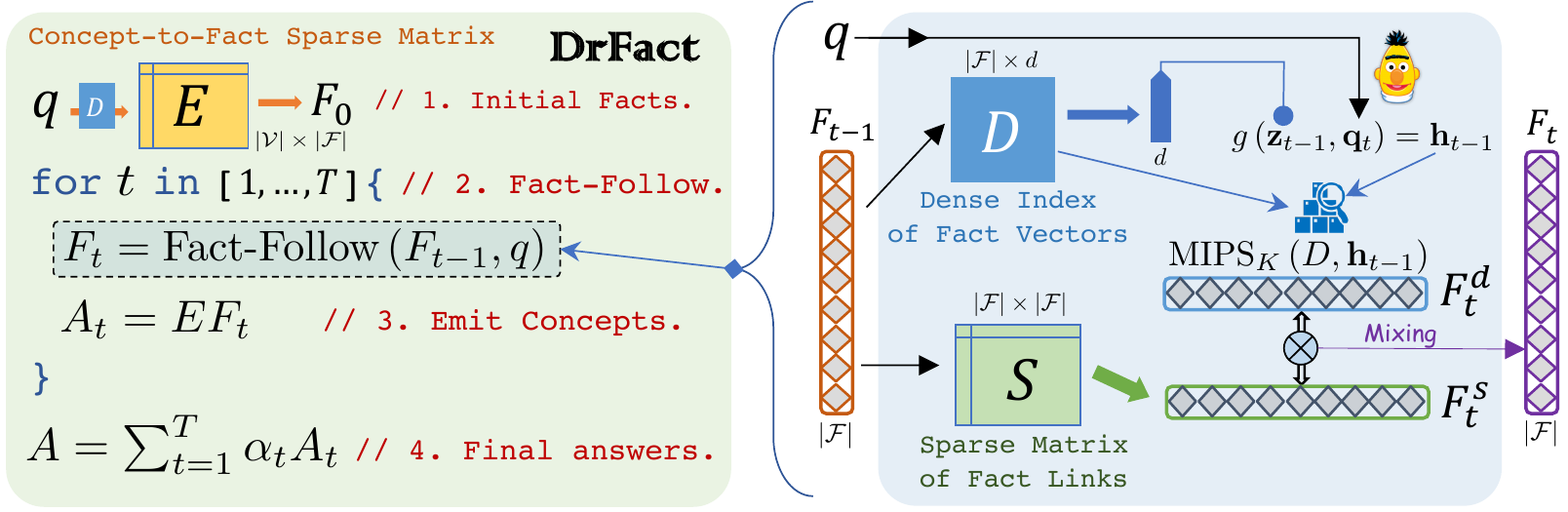}
	\caption{\textbf{The overall workflow of \textsc{DrFact}.} We encode the hypergraph (Fig.~\ref{fig:hypercskg}) with a concept-to-fact sparse matrix $E$ and a fact-to-fact sparse matrix $S$. The dense fact index $D$ is pre-computed with a pre-trained bi-encoder. 
	A weighed set of facts is represented as a sparse vector $F$.
	The workflow (left) of \textsc{DrFact} starts mapping a question to a set of initial facts that have common concepts with it. Then, it recursively performs \texttt{Fact-Follow} operations (right) for computing $F_t$ and $A_t$. Finally, it uses \textit{learnable} hop-weights $\alpha_t$ to aggregate the answers.}
	\label{fig:overview}
\end{figure*}

\subsection{Pre-computed Indices}
\smallskip
\noindent
\textbf{Dense Neural Fact Index $D$.}
We pre-train a bi-encoder architecture over BERT~\cite{Devlin2019}, which learns to maximize the score of facts that contain correct answers to a given question, following the steps of~\citeauthor{dpr} (2020) (i.e., dense passage retrieval), so that we can use MIPS to do dense retrieval over the facts. 
After pre-training, we embed each fact in $\mathcal{F}$ with a dense vector (using the \texttt{[CLS]} token representation). Hence $D$ is a $|\mathcal{F}|\times d$ dense matrix.

\smallskip
\noindent
\textbf{Sparse Fact-to-Fact Index $S$.}
We pre-compute the sparse links between facts by a set of connection rules, such as $f_i \rightarrow f_j$ when $f_i$ and $f_j$ have at least one common concept and $f_j$ introduces at least two more new concepts that are not in $f_i$ (see Appendix~\ref{sec:impl} (2) for more).
Hence $S$ is a binary sparse tensor with the dense shape $|\mathcal{F}| \times |\mathcal{F}|$.

\smallskip
\noindent
\textbf{Sparse Index of Concept-to-Fact Links $E$.}
As shown in Figure~\ref{fig:hypercskg}, a concept can appear in multiple facts and a fact also usually mentions multiple concepts.
We encode these co-occurrences between each fact and its mentioned concepts into a sparse matrix with the dense shape $|\mathcal{V}| \times |\mathcal{F}|$ --- i.e., the \textit{concept-to-fact index}.

\subsection{Differentiable Fact-Following Operation}

The most important part in our framework is how to model the fact-following step in our formulation, i.e., $P\left(F_{t} \mid F_{t-1}, q\right)$. 
For modeling the translation from a fact to another fact under the context of a question $q$, we propose an efficient approach with a differentiable operation that uses both \textit{neural} embeddings of the facts and their \textit{symbolic} connections in the hypergraph.

The symbolic connections between facts are represented by the very sparse fact-to-fact matrix $S$, which in our model is efficiently implemented with the \texttt{tf.RaggedTensor}
construct of TensorFlow \cite{drkit}.
$S$ stores a pre-computed dependency between pairs of facts, $S_{ij}$.
Intuitively, if we can traverse from $f_i$ to $f_j$ these facts should mention some common concepts, and also the facts' semantics are related,
so our $S_{ij}$ will reflect this intuition.
The fact embeddings computed by a pre-trained bi-encoder are in the dense index of fact vectors $D$, which contains rich semantic information about each fact, and helps measure the plausibility of a fact in the context of a given question. 

The proposed fact-follow operation has two parallel sub-steps: 1) sparse retrieval and 2) dense retrieval.
The sparse retrieval uses a fact-to-fact sparse matrix to obtain possible next-hop facts.
We can compute 
$F_{t}^{s} = F_{t-1} S$ 
efficiently thanks to the ragged representation of sparse matrices. 

For the neural dense retrieval, we use a maximum inner product search (MIPS)~\cite{johnson2019billion,scann} over the dense fact embedding index $D$:
\begin{align*}
    \mathbf{z_{t-1}} &= F_{t-1}D\\
    \mathbf{h_{t-1}} &= g(\mathbf{z_{t-1}}, \mathbf{q_t}) \\
    F_{t}^{d} &= \operatorname{MIPS}_K(\mathbf{h_{t-1}}, D)    
\end{align*}
We first aggregate the dense vectors of the facts in $F_{t-1}$ into the dense vector $\mathbf{z_{t-1}}$, which is fed into a neural layer with the query embedding at the current step, $\mathbf{q_t}$ (encoded by BERT), to create a query vector $\mathbf{h_{t-1}}$.
Here $g(\cdot)$ is an MLP that maps the concatenation of the two input vectors to a dense output with the same dimensionality as the fact vectors, which we named to be fact-translating function.
Finally, we retrieve the next-hop top-K facts $F_{t}^d$ with the $\operatorname{MIPS}_K$ operator.

To get the best of both symbolic and neural world, we use element-wise multiplication to combine the sparse and dense retrieved results:
$F_{t} = F_{t}^s \odot F_{t}^d$.
We summarize the fact-following operation with these differentiable steps:
\begin{eqnarray} \label{eq:follow}
F_{t} & = & \operatorname{Fact-Follow}(F_{t-1}, q) \\
  & = & F_{t-1} S \odot \operatorname{MIPS}_K(g(F_{t-1}D, \mathbf{q_t}), D)    \nonumber
\end{eqnarray} 
After each hop, we multiply $F_t$ with a pre-computed fact-to-concept matrix $E$, thus generating $A_t$, a set of concept predictions.
To aggregate the concept scores, we take the maximum score among the facts that mention a concept $c$. Finally
we take the weighted sum of the concept predictions at all hops as the final weighted concept sets $A=\sum_{t=1}^T \alpha_t A_t,$ 
where $\alpha_t$ is a \textit{learnable} parameter. 
{Please read Appendix~\ref{sec:impl} for more details.} 

Equation~\ref{eq:follow} defines a random-walk process on the hypergraph associated with the corpus.  We found that performance was improved by making this a ``lazy'' random walk---in particular by augmenting $F_t$ with the facts in $F_{t-1}$ which have a weight higher than a threshold $\tau$:
$$F_t=  \operatorname{Fact-Follow}(F_{t-1}, q) + \operatorname{Filter}(F_{t-1}, \tau).$$
We call this as \textbf{self-following}, which means that $F_t$ contains highly-relevant facts for all distances $t'<t$, and thus improve models
when there are variable numbers of ``hops'' for different questions.

\textbf{Initial Facts.} 
Note that the set of \textit{initial facts} $F_0$ is computed differently, as they are produced using the input question $q$, instead of a previous-hop $F_{t-1}$.
We first use our pre-trained bi-encoder and the associated index $D$ via MIPS query to finds facts related to $q$, and then select from the retrieved set those facts that contain question concepts (i.e., concepts that are matched in the question text), using the concept-to-fact index $E$.

\subsection{Auxiliary Learning with Distant Evidence}
\label{ssec:aux}
Intermediate evidence, i.e., supporting facts, is significant for guiding multi-hop reasoning models during training.
In a weakly supervised setting, however, we usually do not have ground-truth annotations as they are expensive to obtain.

To get some noisy yet still helpful supporting facts, we use as distant supervision dense retrieval based on the training questions.  Specifically, we concatenate the question and the best candidate answer to build a query to our pre-trained index $D$, and then we divide the results into four groups depending on whether they contain question/answer concepts: 1) question-answer facts, 2) question-only facts, 3) answer-only facts, and 4) none-facts.

Then, to get a 2-hop evidence chain, we first check if a question-only fact can be linked to an answer-only fact through the sparse fact-to-fact matrix $S$.
Similarly, we can also get 3-hop distant evidence.
In this manner,
we can collect
the set of supporting facts at each hop position, denoted as $\{F_1^*, F_2^*, \dots, F_T^*\}$.

The final learning objective is thus to optimize the sum of the cross-entropy loss $l$  between the final weighed set of concepts $A$ and the answer set $A^*$, as well as the auxiliary loss from distant evidence --- i.e., the  mean of the hop-wise loss between the predicted facts $F_t$ and the distant supporting facts at that hop $F_t^*$, defined as follows:
$$\mathcal{L} = {l}(A, A^*) + \frac{1}{T} {\sum_{t=1}^T {l}(F_t, F_t^*)}$$


\section{Experiments}\label{sec:exp}

\subsection{Experimental  Setup}

\subsubsection*{Fact corpus and concept vocabulary }
We use the {GenericsKB-Best} corpus as the main knowledge source\footnote{It was constructed from multiple commonsense knowledge corpora and only kept naturally occurring generic statements, which makes it a perfect fit for OpenCSR.}.
In total, we have {1,025,413} unique facts as our $\mathcal{F}$.
We use the spaCy toolkit
to prepossess all sentences in the corpus and then extract frequent noun chunks within them as our concepts. 
The vocabulary $\mathcal{V}$ has {80,524} concepts, and every concept is mentioned at least 3 times.

\subsubsection*{Datasets for OpenCSR}
To facilitate the research on open-ended commonsense reasoning (OpenCSR),
we reformatted three existing multi-choice question answering datasets to allow evaluating OpenCSR methods.
We choose three datasets:
QASC, OBQA, and ARC, as their questions require commonsense knowledge about science and everyday objects and are presented in natural language.
By applying a set of filters and rephrasing rules,
we selected those open-ended commonsense questions that query concepts in our vocabulary $\mathcal{V}$.

As we know that there can be multiple correct answers for a question in OpenCSR,
we employed crowd-workers 
to collect more answers for each \textit{test} question based on a carefully designed annotation protocol. 
{In total, we collect {15,691} answers for {2,138} rephrased questions for evaluation, which results in 7.5 answers per question on average.}
Please find more details about crowd-sourcing and analysis in Appendix~\ref{sec:opencsrdata}.

We show some statistics of the OpenCSR datasets and our new annotations in Table~\ref{tab:stat}.
To understand the multi-hop nature and the difficulty of each dataset, we use a heuristic 
to estimate the percentage of ``single-hop questions'', for which we can find a fact (from top-1k facts retrieved by BM25) containing both a question concept and an answer concept.
The ARC dataset has about 67\% one-hop questions and thus is the easiest, while OBQA has only 50\%.

\begin{table}[t]
\centering
\scalebox{0.8}{
\begin{tabular}{@{}r|ccc||c}
\toprule
 \textbf{Stat. $\backslash$ Data }    & {ARC}     & {QASC}    & {OBQA}    & \textbf{Overall} \\ \midrule
\# \textbf{All Examples}  &  6,600  & 8,443   &  5,288  & 20,331  \\ \midrule
\# {Training Set} & 5,355   & 6,883   & 4,199   & 16, 437  \\
\# {Validation Set}   & 562     & 731     & 463     & 1,756   \\
\# {Test Set} & 683     & 829     & 626     & 2,138 \\
\midrule
{{Avg.\#Answers}} & 6.8 & 7.6 & 7.7 & 7.5 \\
\midrule
{{Single-hop}}~\% & 66.91\% & 59.35\% & 50.80\% & 59.02\% \\
\bottomrule
\end{tabular}
}
\caption{Statistics of datasets for OpenCSR (v1.0).}
\label{tab:stat}
\end{table}


\subsubsection*{Evaluation metrics.}
Recall that, given a question $q$, the final output of every method is a weighted set of concepts $A=\{(a_1, w_1), \dots \}$. 
We denote the set of \textit{true answer concepts}, as defined above, as $A^*=\{a_1^*, a_2^*, \dots \}$.  
We define \textbf{Hit@K} accuracy to be the fraction of questions for which we can find \textit{at least one} correct answer concept $a_i^*\in A^*$ in the top-$K$ concepts of $A$ (sorted in descending order of weight). 
As questions have multiple correct answers, recall is also an important aspect for evaluating OpenCSR, so we also 
use \textbf{Rec@K} to evaluate the average recall of the top-K proposed answers.
\begin{table*}[t]
\centering
\scalebox{0.9}{
\begin{tabular}{c||cc|cc|cc||cc}
\toprule
 & \multicolumn{2}{c}{ARC} & \multicolumn{2}{c}{QASC} & \multicolumn{2}{c}{OBQA} & \multicolumn{2}{c}{\textbf{\underline{Overall}}} \\ \midrule
 \rowcolor{Gray} Metric  = \textbf{Hit@K (\%)} & H@50  & H@100  & H@50  & H@100  & H@50  & H@100  & H@50  & H@100 \\ \midrule
\cellcolor{cyan!10} BM25 (off-the-shelf)         &  56.95 & 67.35 & 58.50 & 66.71 & 53.99 & 66.29 & 56.48 & 66.78  \\
\cellcolor{cyan!10}DPR~\cite{dpr}        &  68.67 & 78.62 & 69.36 & 78.89 & 62.30 & 73.80 & 66.78 & 77.10  \\
\cellcolor{cyan!10}DrKIT~\cite{drkit}    &  67.63 & 77.89 & 67.49 & 81.63 & 61.74 & 75.92 & 65.62 & 78.48  \\
\cellcolor{cyan!10} \textsc{DrFact} (\textbf{Ours})      &  \textbf{71.60} & \textbf{80.38} & \textbf{72.01} & \textbf{84.56} & \textbf{69.01} & \textbf{80.03} & \textbf{70.87 }& \textbf{81.66 } \\
\midrule
\cellcolor{blue!15} BM25 + MCQA Reranker         &  76.87 & 80.38 & 75.75 & 80.22 & 79.23 & 84.03 & 77.28 & 81.54  \\
\cellcolor{blue!15}DPR + MCQA Reranker           &  76.72 & 83.16 & 81.66 & 87.45 & 77.16 & 83.39 & 78.51 & 84.67  \\
\cellcolor{blue!15}DrKIT + MCQA Reranker         &  78.44 & 83.37 & 84.00 & 86.83 & 79.25 & 84.03 & 80.56 & 84.74  \\
\cellcolor{blue!15}\textsc{DrFact} + MCQA Reranker       &  \textbf{84.19} & \textbf{89.90} & \textbf{89.87} & \textbf{93.00} & \textbf{85.78 }& \textbf{90.10} &\textbf{86.61 }& \textbf{91.00 } \\\midrule \midrule

 \rowcolor{Gray} Metric = \textbf{Rec@K (\%)}  & R@50  & R@100  & R@50  & R@100  & R@50  & R@100  & R@50  & R@100 \\ \midrule 
\cellcolor{cyan!10} BM25 (off-the-shelf)         &  21.12 & 28.08 & 16.33 & 20.13 & 14.27 & 20.21 & 17.24 & 22.81  \\
\cellcolor{cyan!10}DPR~\cite{dpr}        &  28.93 & 38.63 & 23.19 & 32.12 & 18.11 & 26.83 & 23.41 & 32.53  \\
\cellcolor{cyan!10}DrKIT~\cite{drkit}    &  27.57 & 37.29 & 21.25 & 30.93 & 18.18 & 27.10 & 22.33 & 31.77  \\
\cellcolor{cyan!10} \textsc{DrFact} (\textbf{Ours})      &  \textbf{31.48} & \textbf{40.93} & \textbf{23.29} & \textbf{33.60} & \textbf{21.27} & \textbf{30.32} & \textbf{25.35} & \textbf{34.95 } \\
\midrule
\cellcolor{blue!15} BM25 + MCQA Reranker         &  39.11 & 42.96 & 29.03 & 32.11 & 36.38 & 39.46 & 34.84 & 38.18  \\
\cellcolor{blue!15}DPR + MCQA Reranker           &  43.78 & 51.56 & 40.72 & 48.25 & 36.18 & 43.61 & 40.23 & 47.81  \\
\cellcolor{blue!15}DrKIT + MCQA Reranker         &  43.14 & 49.17 & 39.20 & 44.37 & 35.12 & 39.85 & 39.15 & 44.46  \\
\cellcolor{blue!15}\textsc{DrFact} + MCQA Reranker       &  \textbf{47.73 }& \textbf{55.20 }& \textbf{44.30} & \textbf{50.30} & \textbf{39.60} & \textbf{45.24} & \textbf{43.88} &\textbf{50.25}  \\\bottomrule
\end{tabular}
}
\caption{{Results of the \textbf{Hit@K} and \textbf{Rec@K} ($K$=50/100) on OpenCSR (v1.0).
We present two groups of methods with different inference speed levels. The  {\MyColorBox[cyan!10]{upper group}}  is retrieval-only methods that are efficient (\MyColorBox[cyan!10]{$<0.5$ sec/q}), while the {\MyColorBox[blue!15]{bottom group}} are augmented with a computationally expensive answer reranker (\MyColorBox[blue!15]{$\ge 14$ sec/q}).
}}
\label{tab:main}
\end{table*}

\begin{figure}
	\centering
	\vspace{-0.3em}
	\includegraphics[width=1\linewidth]{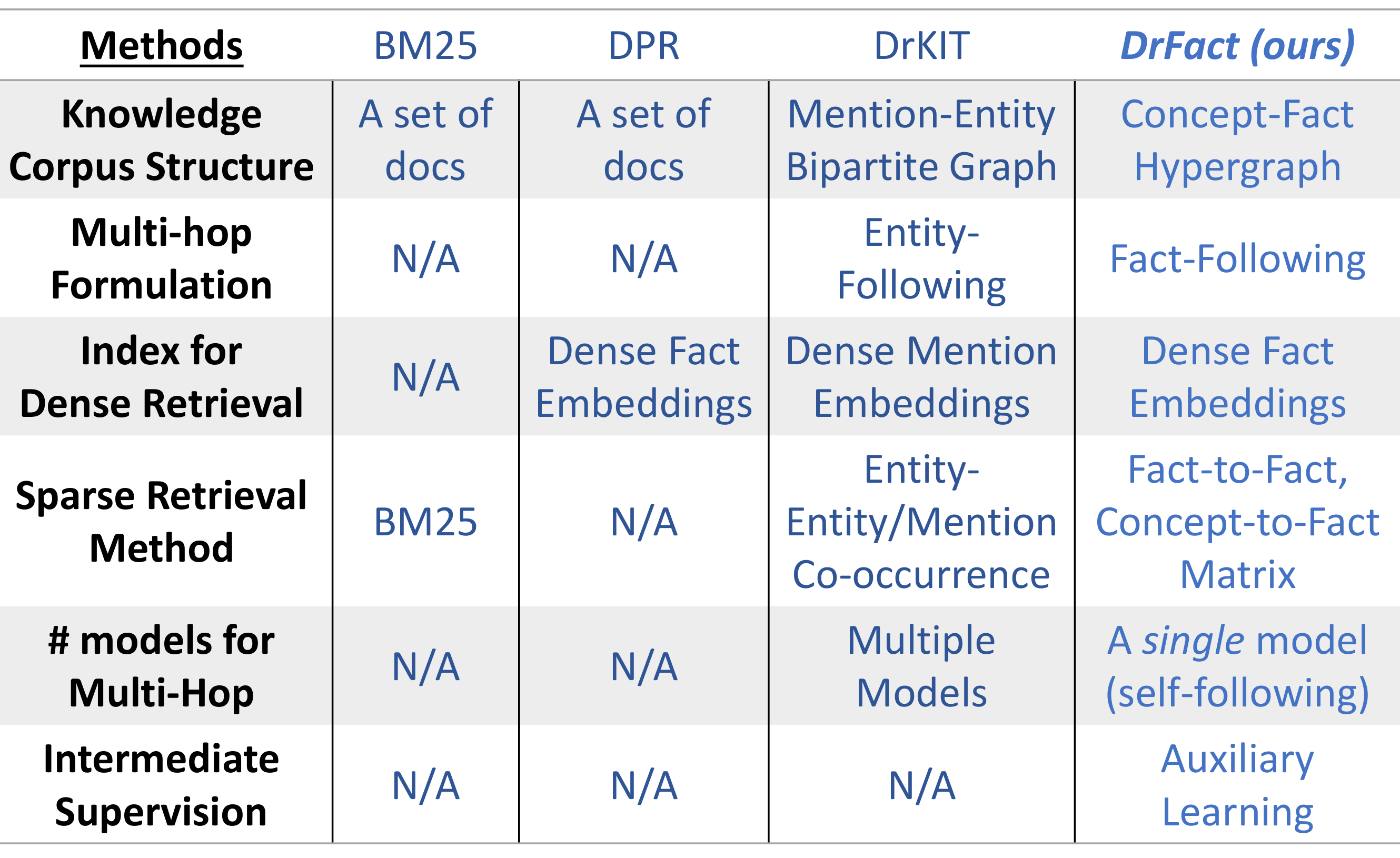}
	\captionof{table}{Comparisons of the four retrieval methods.}
	\label{tab:compare} 
\end{figure}

\subsection{Baseline Methods}  
\label{sec:baseline}
We present baseline methods and an optional re-ranker component for boosting the performance on OpenCSR.
Table~\ref{tab:compare} shows a summary of the comparisions of the three methods and our DrFact.

\smallskip
\noindent
\textbf{Direct Retrieval Methods.}
The most straightforward approach to the OpenCSR task is to directly retrieve relevant facts, and then use the concepts mentioned in the top-ranked facts as answer predictions.
{BM25} is one of the most popular \textit{unsupervised} method for  retrieval, while 
the \textit{Dense Passage Retrieval} ({DPR}) model is a state-of-the-art trainable, neural retriever~\cite{dpr}. 
Following prior work with DPR, 
we used BM25-retrieved facts to create positive and (hard-)negative examples as supervision.
For both methods, we score a concept by the \textit{max}\footnote{We also tried \textit{mean} and \textit{sum}, but \textit{max} performs the best.} of the relevance scores of retrieved facts that mention it. 


\smallskip
\noindent
\textbf{DrKIT.} 
Following~\citeauthor{drkit} (2020), we use DrKIT for OpenCSR, treating concepts as entities. 
DrKIT is also an efficient multi-hop reasoning model that reasons over a pre-computed indexed corpus, which, as noted  above (Sec.~\ref{sec:rel_work}), 
differs from our work in that
DrKIT traverses a graph of entities and entity mentions, while \textsc{DrFact}  traverses a hypergraph of facts.

\smallskip
\noindent
\textbf{Multiple-choice style re-ranking (MCQA).}
A conventional approach to multiple-choice QA (MCQA) is to fine-tune a pre-trained language model such as BERT, by combining a question and a particular concept as a single input sequence in the form of ``\texttt{[CLS]}question\texttt{[SEP]}choice'' and using \texttt{[CLS]} vectors for learning to score choices.
We follow this schema and train\footnote{Specifically, we fine-tune BERT-Large to score truth answers over 9 sampled distractors, and use it to rank the top-500 concepts produced by each above retrieval method.} such a multiple-choice QA model on top of BERT-Large, and use this to re-rank the top-$K$ concept predictions.

\subsection{Results and Analysis}
\noindent
\textbf{Main results.}
For a comprehensive understanding, we report the Hit@K and Rec@K of all methods, at K=50 and K=100, in Table~\ref{tab:main}.
The \textit{overall} results are the average over the three datasets.
We can see that \textsc{DrFact}  outperforms all baseline methods for all datasets and metrics.
Comparing with the state-of-the-art text retriever DPR, \textsc{DrFact}  improves by about 4.1\% absolute points in Hit@50 accuracy overall.
With the expensive yet powerful MCQA reranker module \textsc{DrFact} gives an even large gap ($\sim8\%$ gain in H@50 acc). 

\begin{figure}
	\centering
	\includegraphics[width=1\linewidth]{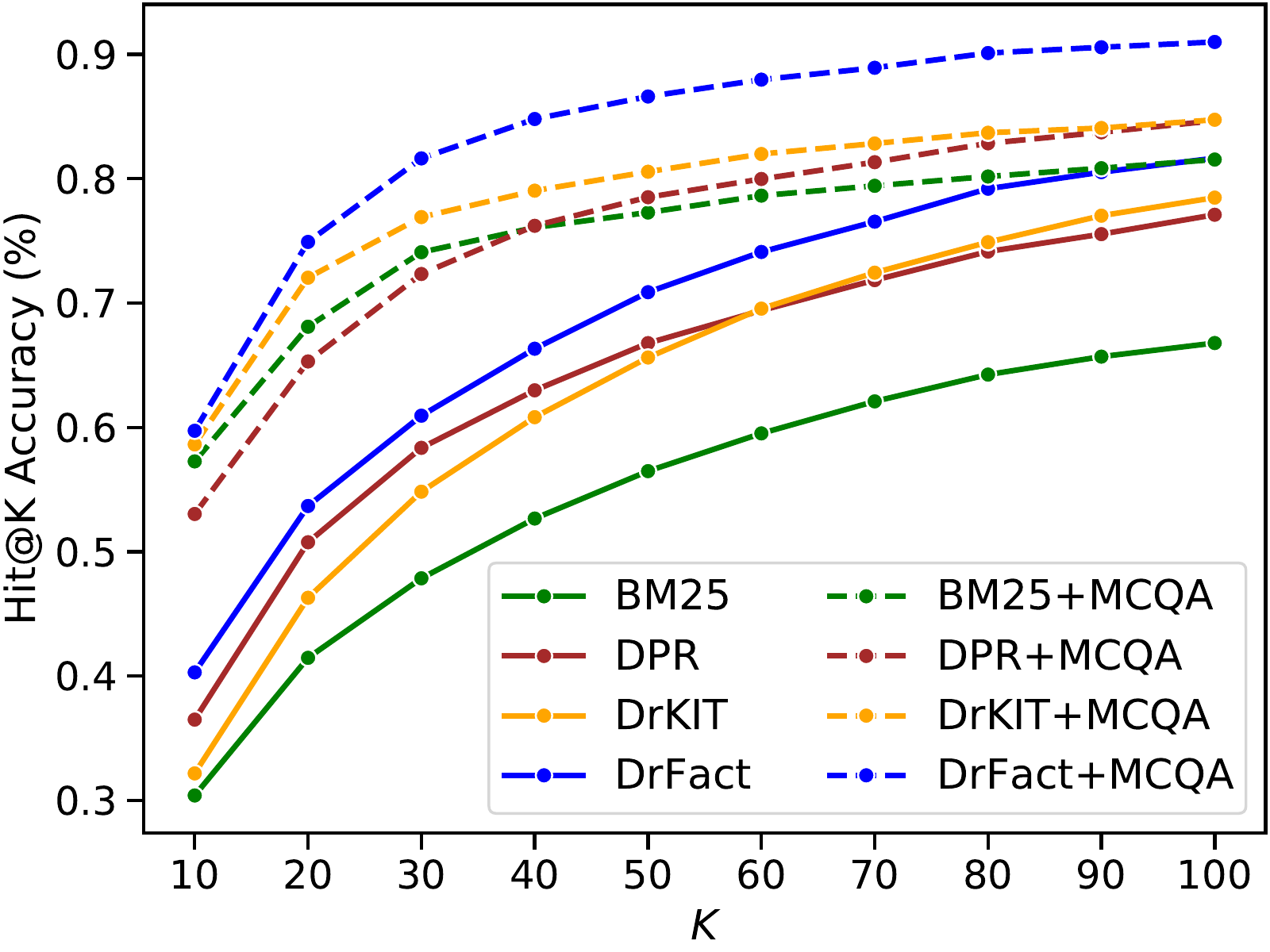}
	\caption{The curve of Hit@K accuracy in \textit{overall}. Please find the curve of Rec@K in Figure~\ref{fig:rkcurve}.}
	\label{fig:hkcurve} 
\end{figure}

The performance gains on the QASC and OBQA datasets are larger than the one on ARC. This observation correlates the statistics that the former two have more multi-hop questions and thus \textsc{DrFact} has more advantages.
As shown in Figure~\ref{fig:hkcurve}, we can see that \textsc{DrFact} consistently outperforms other retrieval methods at different $K$ by a considerable margin.

Interestingly, we find that with the MCQA reranker, DrKIT does not yield a large improvement over DPR, and it usually has a lower than other methods.
We conjecture this is because that entity-centric reasoning schema produces too many possible concepts and thus is more likely to take more irrelevant concepts at the top positions. 

The results on \textbf{Rec@K} in bottom section of Table~\ref{tab:main} show that 
even our \textsc{DrFact}+MCQA model only recalls about 50\% of the correct answers in top-100 results on average. 
This suggests that OpenCSR is still a very challenging problem and future works should focus on improving the ability of ranking \textit{more} correct answers higher.

\smallskip
\noindent
\textbf{Run-time efficiency analysis.}
We use Table~\ref{tab:eff} to summarize the online inference speed of each OpenCSR method.
At inference time, DPR will make one call to BERT-base for encoding a question and do one MIPS search.
Similarly, DrKIT and \textsc{DrFact}  with $T$ hops will make one call to BERT-base for query encoding and do $T$ MIPS searches.
However, since the entity-to-mention matrix ($\operatorname{sp}_{e2m}$) of DrKIT is much larger than the fact-to-fact matrix ($\operatorname{sp}_{f2f}$) of \textsc{DrFact}, DrKIT is about twice as slow as \textsc{DrFact}.
The MCQA is much more computationally expensive, as it makes $K$ calls to BERT-Large for each combination of question and choice. 
Note that in these experiments we use $T$=2 for DrKIT, $T$=3 for \textsc{DrFact} and $K$=500 for the MCQA re-rankers.\footnote{We note the MCQA-reranker could be speed up by scoring more choices in parallel. All run-time tests were performed on NVIDIA V100 (16GB), but MCQA with batch-size of 1 requires only $\sim$5GB.  This suggests more parallel inference on a V100 could obtain 4.5 sec/q for MCQA.}

\begin{table}[]
\centering
\scalebox{0.75
	}{
\begin{tabular}{c|c|c}
\toprule
\textbf{Methods}          & \textbf{Major Computations  }                      & \textbf{Speed} (sec/q) \\ \midrule
\cellcolor{cyan!10}BM25             & Sparse Retrieval                          & 0.14        \\
\cellcolor{cyan!10}DPR              & BERT-base + MIPS                       & 0.08         \\
\cellcolor{cyan!10}DrKIT           & BERT-base + $T$*(MIPS+ $\operatorname{sp}_{e2m}$) & 0.47       \\
\cellcolor{cyan!10}\textsc{DrFact}           & BERT-base + $T$*(MIPS+ $\operatorname{sp}_{f2f}$) & 0.23\\
\midrule
\MyColorBox[cyan!10]{X}+ \MyColorBox[blue!15]{MCQA} & X + $K$ * BERT-Large                        &  + \textit{14.12}    \\ \bottomrule
\end{tabular} 
}
\caption{{The major competitions of each method and their online  (batch-size=1) inference speed in \textit{sec/q}. }}
\label{tab:eff}
\end{table}

\begin{table}[t]
\centering
\scalebox{0.83
	}{
\begin{tabular}{c|ccc||c}
\toprule
                     & ARC     & QASC    & OBQA    & Overall \\ \midrule
$T$=1 ~~~                  & 69.3\% & 70.1\% & 65.0\% & 68.1\% \\
$T$=2  ~~~                 & 71.1\% & 72.2\% & 68.3\% & 70.5\% \\
\rowcolor{Gray}  $T$=3  \cmark  \quad & 71.6\% & 72.0\% & 69.0\% & 70.9\% \\
\midrule
w/o. Self-follow    & 70.9\% & 70.4\% & 68.4\% & 69.9\% \\
w/o.  Aux. loss  & 70.6\% & 70.1\% & 68.0\% & 69.6\% \\ \bottomrule
\end{tabular}
}
\caption{Ablation study of \textsc{DrFact}  (H@50 test acc). }
\label{tab:ablation}
\end{table}

\smallskip
\noindent
\textbf{Ablation study.}
Varying the maximum hops (T=\{1,2,3\}) --- i.e., the number of calls to \texttt{Fact-Follow} --- indicates that overall performance is the best when T=3 as shown in Table~\ref{tab:ablation}. 
The performance with T=2 drops 0.7\% point on OBQA. 
We conjecture this is due to nature of the datasets, in particular the percentage of hard questions.
We also  test the model (with T=3) without the \textit{auxiliary learning loss} (Sec.~\ref{ssec:aux}) or the \textit{self-following} trick.
Both are seen to be important to \textsc{DrFact}.
Self-following is especially helpful for QASC and OBQA, where there are more multi-hop questions.
It also makes learning and inference more faster than an alternative approach of ensembling multiple models with different maximum hops as done in some prior works.

\begin{figure}[t]
	\centering
	\includegraphics[width=1\linewidth]{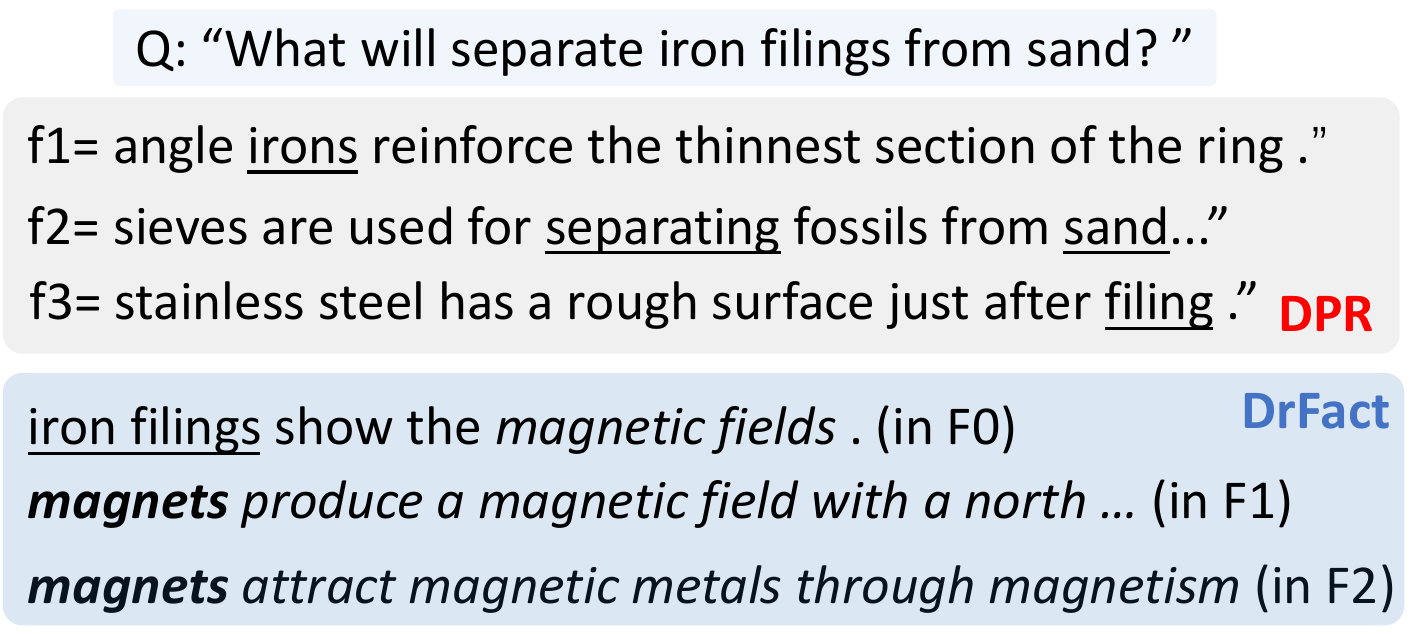}
	\caption{A case study to compare DPR and \textsc{DrFact}.}
	\label{fig:case} 
\end{figure}

\smallskip
\noindent
\textbf{Qualitative analysis.}
We show a concrete example in Fig.~\ref{fig:case} to compare the behaviour of DPR and \textsc{DrFact} in reasoning.
DPR uses purely dense retrieval without any regularization, yielding irrelevant facts.
The fact $f_2$ matches the phrase ``separating...from sand,'' but does not help reason about the question.
The $f_3$ shows here for the semantic relatedness of ``steel'' and ``iron'' while ``filling'' here is not related to question concepts.
Our \textsc{DrFact}, however, can faithfully reason about the question via fact-following over the hypergraph, and use neural fact embeddings to cumulatively reason about a concept, e.g., \textit{magnet}. 
By backtracking with our hypergraph, we can use retrieved facts as explanations for a particular prediction.



\section{Conclusion}\label{sec:conclusion}
We introduce and study a new task --- open-ended commonsense reasoning (OpenCSR) --- which is both realistic and challenging.  
We construct three OpenCSR versions of widely used datasets targeting commonsense reasoning with a novel crowd-sourced collection of multiple answers, 
and evaluate a number of baseline methods for this task.  
We also present a novel method, \textsc{DrFact}.  
\textsc{DrFact} is a scalable multi-hop reasoning method that traverses a corpus (as a hypergraph) via a differentiable  ``fact-following'' reasoning process, employing both a neural dense index of facts and sparse tensors of symbolic links between facts, using a combination of MIPS and sparse-matrix computation. 
\textsc{DrFact} outperforms several strong baseline methods on our data, making a significant step towards adapting commonsense reasoning approaches to more practical applications.
Base on the multi-hop reasoning framework of \textsc{DrFact}, we hope the work can benefit future research on neural-symbolic commonsense reasoning.



\section*{Acknowledgments}
Xiang Ren is supported in part by the Office of the Director of National Intelligence (ODNI), Intelligence Advanced Research Projects Activity (IARPA), via Contract No. 2019-19051600007, the DARPA MCS program under Contract No. N660011924033 with the United States Office Of Naval Research, the Defense Advanced Research Projects Agency with award W911NF-19-20271, and NSF SMA 18-29268.
We thank all reviewers for their constructive feedback and comments.

\section*{*~Ethical Considerations}

\textbf{Crowd-workers.} This work presents three datasets for addressing a new problem, open common-sense reasoning.  The datasets are all derived from existing multiple-choice CSR datasets, and were produced by filtering questions and using crowd-workers to annotate common-sense questions by suggesting additional answers. 
Most of the questions are about elementary science and common knowledge about our physical world.
None of the questions involve sensitive personal opinions or involve personally identifiable information. 
We study posted tasks to be completed by crowd-workers instead of crowd-workers themselves, and we do not retrieve any identifiable private information about a human subject.

\smallskip
\noindent
\textbf{Data bias.} Like most crowdsourced data, and in particular most common-sense data, these crowdsourced answers are inherently subject to bias: for example, a question like ``what do people usually do at work'' might be answered very differently by people from different backgrounds and cultures.  
The prior multiple-choice CSR datasets which our datasets are built on are arguably more strongly biased culturally, as they include a single correct answer and a small number of distractor answers, while our new datasets include many answers considered correct by several annotators.  However, this potential bias (or reduction in bias) has not been systematically measured in this work.

\smallskip
\noindent
\textbf{Sustainability.} 
For most of the experiments,
we use the virtual compute engines on Google Cloud Platform, which ``is committed to purchasing enough renewable energy to match consumption for all of their operations globally.''\footnote{\url{https://cloud.google.com/sustainability}}
With such virtual machine instances, we are able to use the resources only when we have jobs to run, instead of holding them all the time like using physical machines, thus avoiding unnecessary waste.

\smallskip
\noindent
\textbf{Application.} The work also evaluates a few proposed baselines for OpenCSR, and introduced a new model which outperforms them.  
This raises the question of whether harm might arise from applications of OpenCSR---or more generally, since OpenCSR is intended as a step toward making multiple-choice CSR more applicable, 
whether harm might arise more generally from CSR methods.  
Among the risks that need to be considered in any deployment of NLP technology are that responses may be wrong, or biased, in ways that would lead to improperly justified decisions. 
Although in our view the current technology is still relatively immature, and unlikely to be fielded in applications that would cause harm of this sort, 
it is desirable that CSR methods provide audit trails, 
and recourse so that their predictions can be explained to and critiqued by affected parties.  
Our focus on methods that provide chains of evidence is largely a reflection of this perceived need. 

\bibliography{drfact_rebiber} 
\bibliographystyle{acl_natbib}

\clearpage

\appendix

\noindent
{\Large{\textbf{Appendix}}} \\
\smallskip

In this appendix, we show more details of our dataset construction (Appx.~\ref{sec:opencsrdata}), details of model implementation and experiments for reproduciblility (Appx.~\ref{sec:impl}), and more related works (Appx.~\ref{sec:morel}). 
As we have \textit{submitted our code} as supplementary material with detailed instructions for running baselines, we will skip some minor details here. 
We will make our code and data \textit{public} after the anonymity period.

\section{Constructing OpenCSR Datasets}
\label{sec:opencsrdata}
\subsection{Reformatting Questions and Answers}
In this section,
we introduce how we reformat the existing three datasets and crowd-source annotations of multiple answers for evaluating OpenCSR.
To convert a multiple-choice question to an open-ended question, 
we first
remove questions where the correct answer does not contain any concept in $\mathcal{V}$
and the few questions that require comparisons between original choices, as they are designed only for multiple-choice QA, e.g., ``\textit{which} of the following is the \textit{most} \dots''
Then, we rephrase questions with long answers to be an open-ended question querying a single concept.

For example, an original question-answer pair such as (Q:``The Earth revolving around the sun can cause \_\_\_'', A:``{constellation} to appear in one place in spring and another in fall'') is now rephrased to (Q*=``The Earth revolving around the sun can cause \underline{what} to appear in one place in spring and another in fall?'', A*=``constellation'').
Specifically, we combine the  original question (Q) and original correct choice (A) to form a long statement and rephrase it to be a new question (Q*) querying a single concept (A*) in the original answer, where we use the least frequent concept as the target.
This question-rephrasing largely improve the number of answerable questions, particularly for the OBQA dataset.
All are English data.

\subsection{Crowd-sourcing More Answers}
 Note that there can be multiple correct answers to an open-ended question in OpenCSR while the original datasets only provide a single answer.
Thus, we use Amazon Mechanical Turk\footnote{\url{https://www.mturk.com/}} (AMT) to collect more answers for the test questions to have a more precise OpenCSR evaluation.

We design a three-stage annotation protocol as follows:
\begin{itemize} 
    \item S1) {\textbf{Multiple-Choice Sanity Check}}. We provide a question and 4 choices where only one choice is correct and the other 3 are randomly sampled. Only the workers who passed this task, their following annotations will be considered. This is mainly designed for avoiding noise from random workers.
    \item S2) \textbf{Selection from Candidates}. To improve the efficiency of annotation, we take the union of top 20 predictions from BM25, DPR, DrKIT, and DrFact and randomly shuffle the order of these concepts (most of them are about 60$\sim$70 candidates). workers can simply input the ids of the concepts that they think are good answers to the question (i.e., a list of integers separated by comma). There are three different workers for each question and we take the candidates which are selected by at least two workers. Note that we also put the correct answer we already have in the candidates and use them as another sanity check to filter out noisy workers.
    \item S3) \textbf{Web-based Answer Collection}. 
    We generate an URL link to Google Search of the input question to help workers to use the Web for associating more correct answers to the question (the input here is a string for a list of concepts separated by comma). We also provide our concept vocabulary as a web-page so one can quickly check if a concept is valid. 
\end{itemize}

\begin{figure}[t]
	\centering
	\includegraphics[width=1\linewidth]{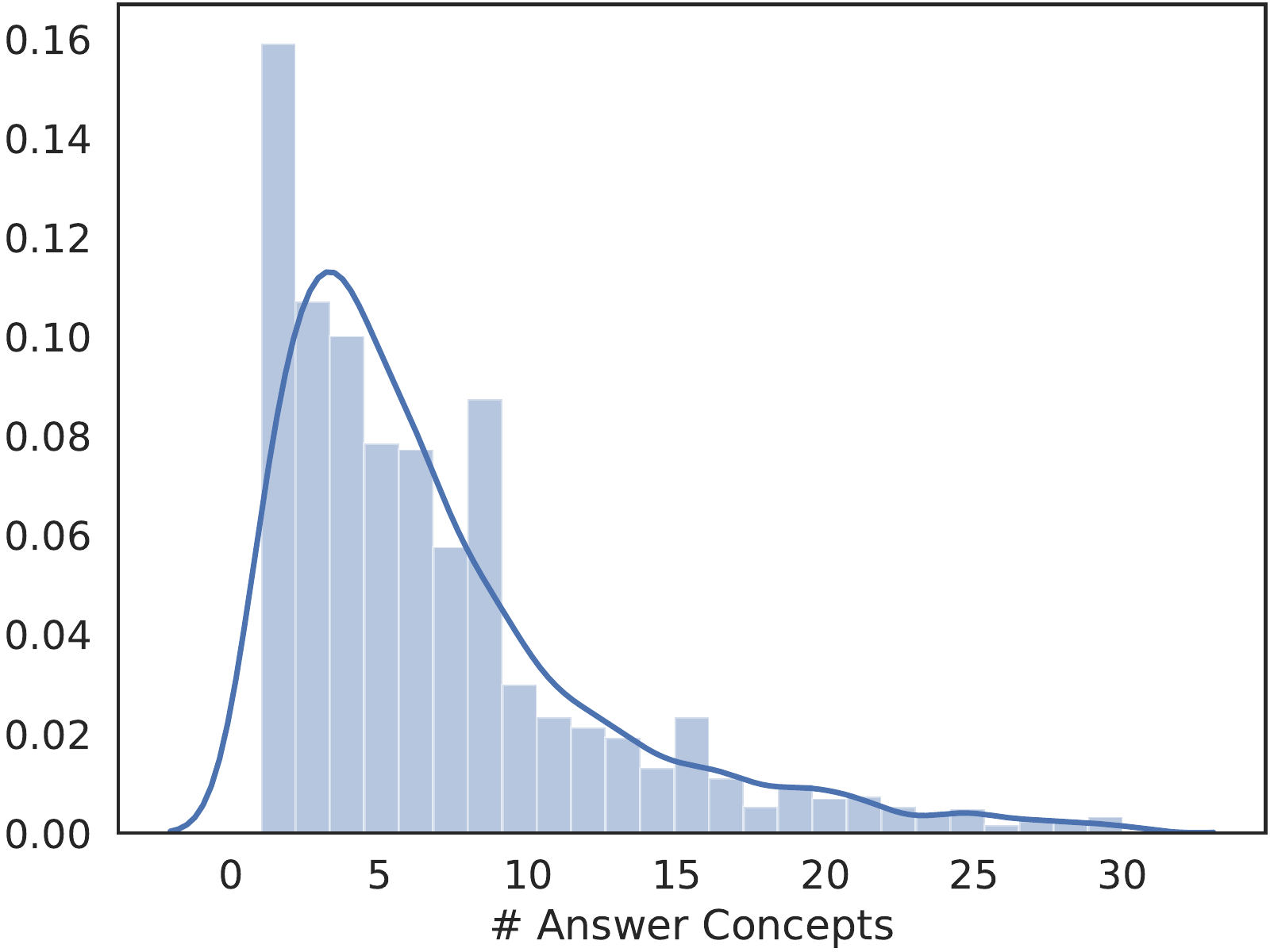}
	\caption{Distribution of \# answers of test questions. \vspace{-1em}}
	\label{fig:answer_dist} 
\end{figure}

After careful post-processing and multiple rounds of re-assignment, we have in total 15k answers for 2k questions, and the distribution of number of answers are in Figure~\ref{fig:answer_dist} and Table~\ref{tab:stat}.

\section{Details of Implementation and Our Experiments}
\label{sec:impl}


\subsection{DrFact Implementation}
We present some concrete design choices within our DrFact implementation which are abstractly illustrated in the main content of the paper.

\smallskip
\noindent
\textbf{(1) Pre-training Dense Fact Index $D$.}
As we mentioned in Sec.~\ref{sec:method},
we follow the steps of~\citeauthor{dpr} (2020) to pre-train a bi-encoder question answering model on top of BERT~\cite{Devlin2019}.
To create negative examples, we use the BM25 results which do not contain any answer concept.
We use BERT-base (\texttt{uncased\_L-12\_H-768\_A-12}) in our implementation and thus $d=768$ in our experiments.

\smallskip
\noindent
\textbf{(2) Sparse Fact-to-Fact Index $S$.}
We use a set of rules to decide if we can create a link $f_i \rightarrow f_j$ (i.e., $S_{ij}=1$) as follows:
\begin{itemize} 
    \item $i\neq j$. We do not allow self-link here but use \textit{self-following} as we described in Sec.~\ref{sec:method}.
    \item $|I|>=1$ where $I$ is the set of concepts that are mentioned in both $f_i$ and $f_j$. Note that we remove the most frequent 100 concepts (e.g., human) from $I$.
    \item $|I| < |f_i|$. We do not create links when all concepts in $f_i$ are mentioned in $f_j$, which are usually redundant.
    \item $|f_j| - |I| >=2$. We create links only when there are more than two unseen concepts in $f_j$ which are not in $f_i$, such that the fact-to-fact links create effective reasoning chains.
\end{itemize}

We also limit that a fact can be followed by at most 1k different facts. Additionally, we append the links from our distant supervision of justifications as well if they were filtered out before.

\smallskip
\noindent
\textbf{(3) Hop-wise Question Encoding $\mathbf{q_t}$.}
We encode the question $q$ with BERT-base and then use its \texttt{[CLS]} token vector as the dense representation for $\mathbf{q}$.
For each hop, we append a hop-specific layer to model how the question context changes over the reasoning process --- 
$\mathbf{q_t} = \operatorname{MLP}_{\theta_t}(\mathbf{q})$.

\smallskip
\noindent
\textbf{(4) Fact Translating Function $g$.}
The translating function accepts both the vector representation of previous-hop facts $\mathbf{F_{t-1}}$ and the hop-wise question vector $\mathbf{q_t}$ and uses an MLP to map the concatenation of them to a vector used for a MIPS query:
$\mathbf{h_{t-1}}=\operatorname{MLP}_{\theta_g}([\mathbf{F_{t-1}};\mathbf{q_{t}}])$.
Thus, $\mathbf{h_{t-1}}$ has the same dimension as a fact vector in $U$.

\smallskip
\noindent
\textbf{(5) Hop-wise Answer Weights $\alpha_t$.}
We use the shared query vector to learn how to aggregate predictions at different hops.
For a $T$-hop DrFact model, we learn to transform the $\mathbf{q}$ to a $T$-dim vector where $\alpha_t$ is the $t$-th component.


\begin{figure}
	\centering
	\hspace{-2em}
	\includegraphics[width=1\linewidth]{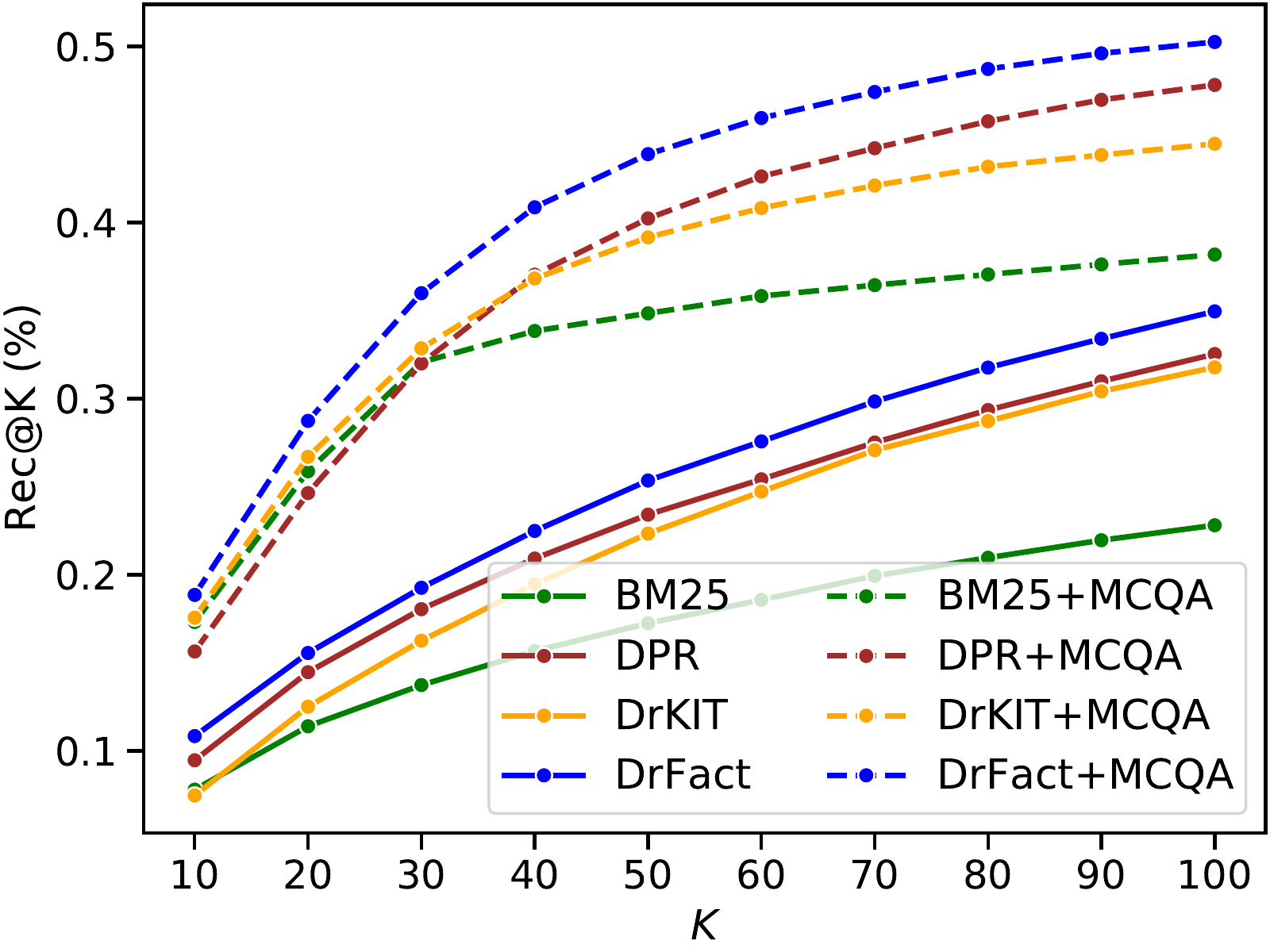}
	\caption{The curve of Rec@K in \textit{overall} data.}
	\label{fig:rkcurve} 
\end{figure}
\subsection{Hyper-parameters and Training Details}
\label{sec:hp}
We now present the details and final hyper-parameters that we used in our experiments.
For all methods, we tune their hyper-parameters on the validation set and then use the same configurations to train them with the combination of the training and validation sets for the same steps. 

\noindent
\textbf{BM25.} We use the off-the-shelf implementation by elasticsearch\footnote{\url{https://github.com/elastic/elasticsearch}}, which are open-source and unsupervised.
For the run-time analysis, we use Intel(R) Xeon(R) CPU @ 2.00GHz and the localhost webserver for data transfer.

\smallskip
\noindent
\textbf{DPR.} We use the source code\footnote{\url{https://github.com/facebookresearch/DPR}} released by the original authors. The creation of negative contexts are the same when we pre-train our dense fact index $D$, which are sampled from BM25 results.

\smallskip
\noindent
\textbf{DrKIT.}
We use the official source code\footnote{\url{https://github.com/google-research/language/tree/master/language/labs/drkit}} for our experiments. 
We did minimal modifications on their code for adapt DrKIT towards building dense index of mentions for the OpenCSR corpus and datasets. For fair comparisions between DPR, DrKIT and DrFact, we all use BERT-base as question and mention/fact encoder. We use 200 as the dimension of mention embeddings and T=2 as the maximum hops. We found that using T=3 will cause too much memory usage (due to denser entity-to-mention matrix) and also result in a very slow training speed. Non-default hyper-parameters are: \textit{train\_batch\_size}=8 due to the limit of our GPU memory, \textit{entity\_score\_threshold}=5e-3 (out of \{5e-2, 5e-3, 5e-4, 1e-4\}) to filter numerous long-tail intermediate concepts for speeding up training and inference. 

\smallskip
\noindent
\textbf{DrFact.}
Similar to DrKIT, we also implement DrFact in TensorFlow for its efficient implementation of \texttt{tf.RaggedTensor} which are essential for us to compute over large sparse tensors. 
We record the default hyper-parameters in our submitted code. 
We use a single V100 GPU (16GB) for training with batch size of 24 (using 15GB memory) and learning rate as 3e-5, selected from \{1e-5, 2e-5, 3e-5, 4e-5, 5e-5\}. 
The \textit{entity\_score\_threshold}=1e-4, and \textit{fact\_score\_threshold}=1e-5, which are all selected from \{1e-3, 1e-4, 1e-5\} based on the dev set.

\smallskip
\noindent
\textbf{Model Parameters.}
DPR, DrKIT and DrFact are all based on the BERT-base, which are 110 million parameters (after pre-training index).
DrKIT and DrFact additionally have several MLP layers on top of `[CLS]' token vectors, which are all less than 1 million parameters.
The MCQA-reranker model is based on BERT-Large, and thus has 345 million parameters.

\section{Discussion on Other Related Work}
\label{sec:morel}

\smallskip
\noindent
\textbf{Other Open-Domain QA models.}
Recent open-domain QA models such as REALM~\cite{guu2020realm}, Path-Retriever~\cite{asai2020learning}, ORQA~\cite{lee2019latent}, and RAG~\cite{lewis2020retrieval}, 
mainly focus on QA over the full Wikipedia corpus like DrKIT~\cite{drkit} does. 
Some of them explicitly use the links between pages to form reasoning chain, while a few them rely on expensive QA-oriented pre-training. 
Moreover, as DPR~\cite{dpr} already shows better performance (see their Table 4) than most prior works with a simpler method, we thus use DPR as the major baseline for evaluation in this work.

\end{document}